\documentclass[letterpaper, 10 pt, conference]{ieeeconf} 
\IEEEoverridecommandlockouts
\usepackage{graphicx}
\usepackage{subfig}
\usepackage{amsmath}
\usepackage{mathtools}
\usepackage{bm}
\usepackage{esvect}
\usepackage{gensymb}
\usepackage{amsmath}
\usepackage{amssymb}
\usepackage{subfig}

\title{\LARGE \bf
Explicit Motion Risk Representation
}

\author{Xuesu Xiao$^{1}$, Jan Dufek$^{1}$, and Robin Murphy$^{1}$% <-this % stops a space
\thanks{$^{1}$Xuesu Xiao, Jan Dufek, and Robin Murphy are with the Department of Computer Science and Engineering,
        Texas A\&M University, College Station, TX 77843
        {\tt\small \{xiaoxuesu, dufek, robin.r.murphy\}@tamu.edu}}%
}

\begin{document}

\maketitle
\thispagestyle{empty}
\pagestyle{empty}

%%%%%%%%%%%%%%%%%%%%%%%%%%%%%%%%%%%%%%%%%%%%%%%%%%%%%%%%%%%%%%%%%%%%%%%%%%%%%%%%
\begin{abstract}
% objectives with no jargon
This paper presents a formal definition and explicit representation of robot motion risk. 
% how it is done today
Currently, robot motion risk has not been formally defined, but has already been used in motion and path planning. Risk is either implicitly represented as model uncertainty using probabilistic approaches, where the definition of risk is somewhat avoided, or explicitly modeled as a simple function of states, without a formal definition. 
% what is new
In this work, we provide formal reasoning behind what risk is for robot motion and propose a formal definition of risk in terms of a sequence of motion, namely path. Mathematical approaches to represent motion risk are also presented, which is in accordance with our risk definition and properties. 
% Who cares
The definition and representation of risk provide a meaningful way to evaluate or construct robot motion or path plans. The understanding of risk is even of greater interest for the search and rescue community: the deconstructed environments cast extra risk onto the robot, since they are working under extreme conditions. A proper risk representation has the potential to reduce robot failure in the field. 

\end{abstract}

%%%%%%%%%%%%%%%%%%%%%%%%%%%%%%%%%%%%%%%%%%%%%%%%%%%%%%%%%%%%%%%%%%%%%%%%%%%%%%%%
\section{INTRODUCTION}
\label{sec::introduction}
Safety, security, and rescue applications are example domains where robots are used to substitute human agents to undertake risk in dangerous, dirty, and dull (DDD) environments \cite{tiwari2019unified}. While guaranteeing safety by projecting human presence, the robots must also ensure their own safety at all times, in lieu of mission-critical or economic considerations. This becomes challenging in unstructured or confined spaces, such as after-disaster deployments, when robots need to work under extreme conditions. Therefore, it is necessary for the robots to locomote in a risk-aware manner in those deconstructed environments to maximize the possibility of safe mission completion. 

Mission risk exists due to a variety of reasons, e.g., structural collapse after an earthquake or flooding after a major hurricane. However, a significant portion of risk during mission execution under extreme conditions is caused by the locomotion of the robot, from both the robot's internal components and external interaction with the deconstructed environments. During mission planning and execution, risk caused by motion could be actively controlled and mitigated by the robot, unlike other risk sources. Therefore reasoning about motion risk could help the robot to conduct safe and trust-worthy motion in those challenging scenarios. 

\begin{figure}
\centering
\subfloat[]{\includegraphics[width=0.4\columnwidth]{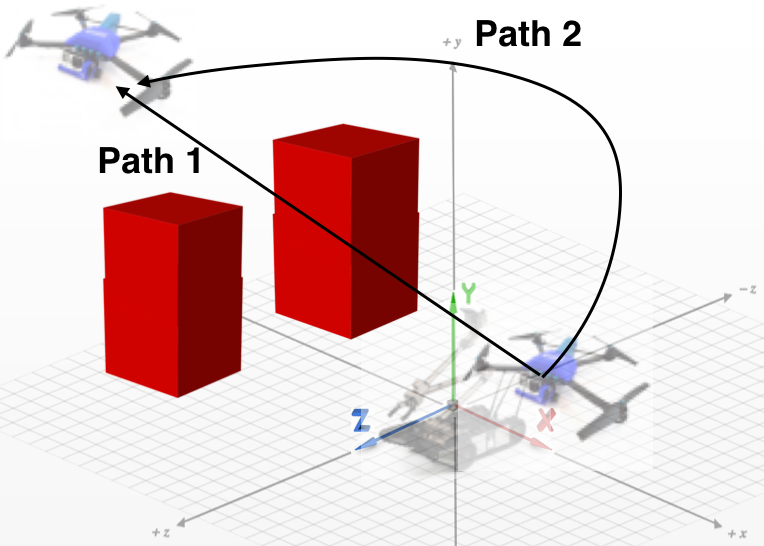}%
\label{fig::va}}
\hfill
\subfloat[]{\includegraphics[width=0.6\columnwidth]{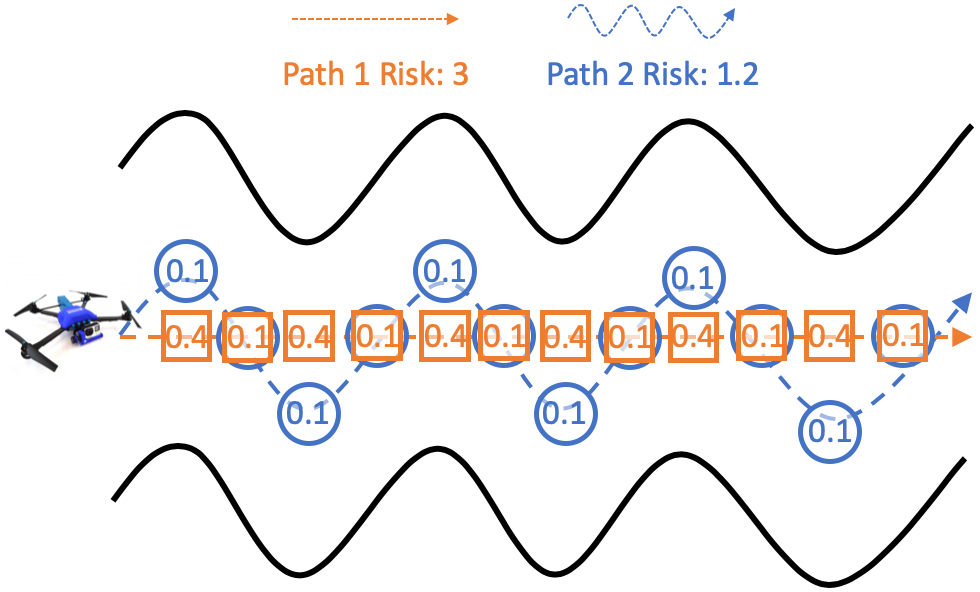}%
\label{fig::tortuous_straight}}
\caption{Motion Risk in Terms of Path: executing a sequence of motion, namely path, inherently entails taking risk. Although starting and ending at same locations, taking different paths put the robot at different risk levels. }
\label{fig::path_examples}
\end{figure}

However, to our best knowledge, a formal definition of motion risk does not exist. Although risk-aware planning has assumed risk in the form of various cost functions or probabilistic sensor or action models, a formal definition and a general representation of motion risk is still missing. In this work, we provide formal reasoning behind robot motion risk and propose a formal definition of risk in terms of a sequence of motion, i.e. path (Fig. \ref{fig::path_examples}), along with general mathematical approaches to explicitly represent motion risk. Based on our risk definition and representation, different robot paths could be evaluated and compared using the same metric. Planners can use this general risk representation to plan risk-aware or risk-averse path, which could reduce robot mission failure, especially in deconstructed and extreme work envelope. 

The rest of the paper is organized as follows: Sec. \ref{sec::related_work} gives related work about robot motion risk in the literature. Sec. \ref{sec::representation} formally defines and explicitly represents motion risk in a general and comprehensive way. Sec. \ref{sec::example} presents quantitative motion risk representation results on a particular robot platform and visualizes path examples with their corresponding risk. Discussions on the risk definition and representation are also provided. Sec. \ref{sec::conclusions} concludes the paper. 

\section{RELATED WORK}
\label{sec::related_work}

Although researchers have investigated risk-aware planning from a variety of directions, including Partially Observable Markov Decision Process (POMDP) with negative reward as risk penalty \cite{pereira2013risk}, constrained POMDP \cite{undurti2010online}, or risk allocation \cite{ono2008iterative}, the definition of risk was always taken for granted and only based on the ad hoc applications. 

A large body of work implicitly treated risk as model uncertainty. Risk was not explicitly defined or represented, but embedded in certain probabilistic models. The rationale behind this was that in a perfectly known world where the robot sensor and transition model is deterministic, the robot is not facing any motion risk at all. With this implicit risk representation, the following planning was conducted in belief space \cite{bry2011rapidly, van2012motion, agha2014firm}. In this approach, however, the definition of risk is avoided using probabilistic models, which is difficult to convincingly acquire for field robotics. The hidden risk definition and representation is neither illustrative nor intuitive for reasoning. 

Researchers have looked at explicit representation of risk as well, although a formal definition of risk is still lacking. \cite{gu2006comprehensive} proposed risk as an accumulative parameterized function based on distance to threats. A similar approach was taken by \cite{de2011minimum}, where a risk map was generated based on ground orography and the risk of each location could be looked up from the map. Those works based risk only on the distance to some ad hoc hazards. \cite{soltani2004fuzzy} further represented risk of navigating in construction sites in terms of not only distance to hazard zones but also visibility value as a function of state. All those approaches treated risk as a function of state (or location) on the path, and the total motion risk of executing that path was simply the summation of all states' risk (Fig. \ref{fig::tortuous_straight}). This only focused on one very specific form of risk, neglecting other more general forms. 

It seems reasonable that risk associated with each individual step should be fully embedded in the state. However, it overlooked the fact that the path which are composed of states in terms of a sequence of Cartesian waypoints is only a projection of the true motion trajectory in a lower dimensional space, i.e., higher order derivatives are neglected. More specifically, given any type of locomotive system, its true motion trajectory could be expressed by its dynamic model in the state space.

\begin{equation}
\label{eqn::discrete_state_space2}
\left\{\begin{matrix}
\bm{x}_{t+1}=\bm{f}(t, \bm{x}_{t}, \bm{u}_{t})\\
\bm{z}_t=\bm{g}(t, \bm{x}_{t}, \bm{u}_{t})
\end{matrix}\right.
\end{equation}

%\begin{equation}
%\label{eqn::cont_state_space1}
%\left\{\begin{matrix}
%\bm{\dot{x}}(t)=\bm{f}(t, \bm{x}(t), \bm{u}(t))\\
%\bm{z}(t)=\bm{g}(t, \bm{x}(t), \bm{u}(t))
%\end{matrix}\right.
%\end{equation}

where $\bm{x}_t \in \mathbb{R}^n$, $\bm{z}_t \in \mathbb{R}^m$, $\bm{u}_t \in \mathbb{R}^p$ are the state vector, output vector, and input vector, respectively. The functions $\bm{f}(\cdot, \cdot, \cdot)$ and $\bm{g}(\cdot, \cdot, \cdot)$ describe the system and output dynamics over time. For the scope of this research, we focus on discrete locomotion systems and it is assumed that the state vector $\bm{x}_t$ is directly observable, therefore the output vector $\bm{z}_t$ is ignored. Ideally, the state vector $\bm{x}$ captures necessary information at time step $t$ to determine the risk the locomotive system is facing at this point, as a function $R: \bm{x} \mapsto ri$, where $ri \in \mathbb{R}$ is the risk index for that particular state vector. 

However, in robotic path and motion planning, planning in the full dimensional state space is not always computationally feasible. Therefore the controls over higher order of derivatives of the state vector is usually assumed to be handled separately in the form of low level feedback or feedforward based controllers: the state of the robot  of interest is hence represented in a much lower dimensional space ($<< n$). For example, path planning assumes the full trajectory is only in a three dimensional Cartesian space and the acceleration or even the velocity are ignored in the planner, assuming the low-level controllers are able to drive those high order derivatives into their desired values. This practical approach for path or motion planning by projecting the full dimensional state space into a computationally tractable sub-space, however, ignores the potential information for planning from those reduced dimensions, e.g., executing those high order derivatives may inherently entail taking risk or other system states could also introduce risk. The practical dimensionality reduction techniques exclude the possibility of these information being considered for further purposes, such as risk representation or planning. 

This work firstly presents a formal definition of motion risk in general. It still works on the reduced state space, but takes into account the entire path leading to a state, instead of only the state by itself, to incorporate as much missing information as possible into a more comprehensive risk representation. 

%$\bm{\dot{x}}(t)$ is the derivative of $\bm{x}(t)$:

%\begin{equation}
%\label{eqn::derivative}
%\bm{\dot{x}}(t) := \frac{d}{dt}\bm{x}(t)
%\end{equation}

%It starts with the preliminaries of general dynamic systems and an ideal full dimensional state-dependent risk representation. Then the focus is narrowed down to practical state representation in robotic path and motion planning in a lower dimensional space, and points out the missing history information given only one single state in the projected state space. Therefore, our proposed explicit risk representation needs to incorporate ``path-level'' information in order to compensate the missing dimensionality in our reduced state space. 

\section{EXPLICIT RISK REPRESENTATION}
\label{sec::representation}
This section presents the explicit risk representation proposed in this research. It starts with a formal definition of risk in terms of path as in a practical low dimensional space for robotic path and motion planning. Due to the missing history information given only one single state in the reduced state space, we present approaches to augment the low dimensional path and incorporate ``path-level'' information in order to compensate the missing dimensionality. Then risk caused by different aspects of the augmented space is represented in a quantitative manner, which observes the formal definition of risk. 

\subsection{Risk Definition}
Risk is one embodiment of uncertainty. We propose one possible way of defining risk in terms of a sequence of motion, namely path: risk is  the \emph{relative likelihood} of the robot \emph{not being able to finish the path}. It is a relative measurement of certain relevant features of paths with respect to a certain robot. The two components of the definition will be formally defined as well. 

\subsubsection{\emph{Relative Likelihood}}
In this definition, \emph{Relative Likelihood} gives the relative order of the likelihood of some event happening, e.g., the robot not being able to finish the path. The order is reflected by a numerical value proportional to the likelihood. This numerical value is called \emph{risk index}. This \emph{risk index} is to be distinguished from other uncertainty measurements such as probability or possibility, as they already have their own strict mathematical definitions. Our \emph{risk index} only gives the relative order of the likelihood: a higher \emph{risk index} only means it is more likely that the robot cannot finish the path than a lower \emph{risk index}, but no conclusions or comparison about the probability or possibility of failure could be drawn. 

\subsubsection{\emph{Not Being Able To Finish The Path}}
We also need to formally define \emph{not being able to finish the path}. In this work, we focus on discrete state spaces. For a discrete reduced state space, a feasible and collision-free path could be represented in the form of an ordered sequence of states:

\begin{equation}
\label{eqn::path}
P = \{s_0, s_1, ..., s_n\}
\end{equation}

where $s_i \in \mathbb{R}^3$ (or $\mathbb{R}^2$ if the robot resides in 2-D workspace) denotes the $i$th state on path $P$ while $s_0$ is the initial state. For the path to be feasible, two consecutive states need to be locally connected, which is to satisfy 

\begin{equation}
\label{eqn::feasible}
\lVert s_i - s_{i-1}\rVert_2 \leq r_c, \forall 1 \leq i \leq n
\end{equation}

Here $r_c$ is the radius of connectivity, ascribing the maximum distance between two consecutive states for feasibility. The robot's workspace is also occupied with a set of obstacles:

\begin{equation}
\label{eqn::obstacles}
\mathbb{OB}= \{ob_i | i = 1, 2, ..., o\}
\end{equation}

where $ob_i \in \mathbb{R}^3$ has radius $r_{{ob}_i}$. For the path $P$ to be collision-free:

\begin{equation}
\label{eqn::collision-free}
\forall 1\leq i\leq n, 1\leq j\leq o, \lVert s_i - ob_j\rVert_{N}\geq r_{{ob}_j}
\end{equation}

Here, the $N$-norm could be $1$, $2$, $\infty$, etc., depending on the representation of the obstacle. Given a workspace, the obstacle set $\mathbb{OB}$ is assumed to be constant and therefore when being used as an argument of a function, $\mathbb{OB}$ is omitted. 

The execution $E$ of the planned path $P$ is returned at a terminal state, either reported by the robot after completion or due to system failure such as a crash, described by another ordered sequence of actually executed states: 

\begin{equation}
\label{eqn::execution}
E = \{e_0, e_1, ..., e_m\}
\end{equation}

where $e_i \in \mathbb{R}^3$, $e_0 = s_0$ since the path plan should starts at the robot's initial position. We firstly define \emph{being able to finish the path}, whose negation is simply \emph{not being able to finish the path}: a robot is able to finish the path plan $P$ with path execution $E$ if 

%$\forall 0 \leq i_1 \leq n$, we have 

%\begin{equation}
%\label{eqn::finishes1}
%\begin{split}
% &\exists 0 \leq j_1 \leq m, \lVert s_{i_1} - e_{j_1} \rVert_2 \leq r_p\\
%\land &\forall i_2 \geq i_1, \exists j_1 \leq j_2 \leq m, \lVert s_{i_2} - e_{j_2} \rVert_2 \leq r_p
%\end{split}
%\end{equation}

\begin{equation}
\label{eqn::finishes1}
\begin{split}
\forall 0 \leq i < n, ~~&\exists 0 \leq j_1 \leq m, \lVert s_i - e_{j_1} \rVert_2 \leq r_p\\
\land &\exists j_1 \leq j_2 \leq m, \lVert s_{i+1} - e_{j_2} \rVert_2 \leq r_p
\end{split}
\end{equation}

The first part of Eqn. \ref{eqn::finishes1} guarantees that for all positions on the path plan, there exists at least one position in the actual execution that is within $r_p$ distance. This makes sure that all states on the path plan is reached by the robot. The second part after the \emph{AND} operation of Eqn. \ref{eqn::finishes1} guarantees the chronological order of the states.  

However, a random-walk-like execution in the free space will possibly satisfy Eqn. \ref{eqn::finishes1} with arbitrarily long execution steps. So its symmetric condition, Eqn. \ref{eqn::finishes2} is added in order to guarantee the actual execution cannot be too far away from the plan:

\begin{equation}
\label{eqn::finishes2}
\begin{split}
\forall 0 \leq j < m, ~~&\exists 0 \leq i_1 \leq n, \lVert e_j - s_{i_1} \rVert_2 \leq r_e\\
\land &\exists i_1 \leq i_2 \leq n, \lVert e_{j+1} - s_{i_2} \rVert_2 \leq r_e
\end{split}
\end{equation}

Usually $r_e \geq r_p$ so that the execution can deviates more from the path than vice versa. We name Eqn. \ref{eqn::finishes1} \emph{reachability condition} and  Eqn. \ref{eqn::finishes2} \emph{stability condition}. 

If the robot's execution $E$ of plan $P$ satisfies both \emph{reachability condition} (Eqn. \ref{eqn::finishes1}) and \emph{stability condition} (Eqn. \ref{eqn::finishes2}), it is able to finish the path. Therefore, violation of any of the conditions is defined as \emph{not being able to finish the path}. For violating \emph{reachability condition}, it is possiblly because of inappropriate interactions with the environment, e.g., the robot collides with an obstacle and cannot continue with path execution, or because the robot's internal components fail due to frequent aggressive actions along the path. So the last executed position $e_m$ only corresponds to some middle state in $P$, leaving following states not reached at all and therefore violating \emph{reachability condition}. Another scenario is that only some middle states $s_i$ in $P$ are not reached by any $e_j$ in $E$. But this is less likely since the robot's low level navigator or controller usually makes sure that the robot reaches a certain state $s_i$ before it moves on to $s_{i+1}$. Violating \emph{stability condition} can be viewed as certain portion of the execution deviates too much from the plan, e.g., due to sharp turns, overshoot, or disturbances. If this is not the concern, $r_e$ could be set to $\infty$ so that Eqn. \ref{eqn::finishes2} becomes trivial and the robot finishes the path as long as all states $s_i$ on the plan $P$ is reached by some $e_j$ in execution $E$. 

It is worth to note that this work addresses the motion risk with respect to a path plan $P$, in order to predict the relative likelihood that a a future execution $E$ fails the path. With this formal definition of risk, the \emph{risk index} used has the following properties: 

\begin{itemize}
\item Non-negativity 
\item Monotonicity 
\item Additivity
\end{itemize}

As a measurement of relative likelihood, \emph{risk index} should be non-negative. With increasing states in the path plan $P$, \emph{risk index} is monotonically increasing. It is trivial to show that the risk of executing a path plan $P$ contains the risk of executing its sub-path, and the execution of the rest of the path will induce extra risk due to non-negativity. Given a path $P$ composed of two disjoint path segments $P_1$ and $P_2$, the likelihood of not being able to finish the whole path could be interpreted as the addition of the likelihood of both segments: $ri(P_1\cup P_2)=ri(P_1) + ri(P_2)$ if $P_1$ and $P_2$ are disjoint. 

\subsection{Augmented Lower Dimensional State}
In a conventional state-dependent risk representation, risk at state $s_i$ is defined based on a function mapping from one state to a risk index $r: s_i \mapsto ri$ and the risk of executing a path $P$ is a simple summation of all individual states $risk(P) = \sum_{i=0}^{n} r(s_i)$. For a more general and comprehensive risk representation, the function $r(\cdot)$ is either not well-defined (due to the lack of history information) or only contains a subset of all risk elements (this is simply the subset of risk elements which are state-dependent). This is the reason why the risk information enclosed in the high order derivatives in the original full dimensional space is missing. 

In the proposed path-dependent risk representation, risk at state $s_i$ cannot be simply evaluated by the state alone, but also by the path leading to $s_i$, $p_i = \{s_0, s_1, ..., s_i\}$. The risk at $s_i$ is computed through the mapping $R: (s_0, s_1, ..., s_i) \mapsto ri$. The path-level risk is relaxed from the summation of state-dependent risk only $risk(P) = \sum_{i=0}^{n} r(s_i)$ to a more general representation $risk(P) =  \sum_{i=0}^{n} R(p_i)$, therefore partially recovers the missing information due to dimensionality reduction. 

In order to do so, we further augment the reduced \emph{state} with \emph{action} and \emph{path}. \emph{action} is defined as the transition between two consecutive states: 

\begin{equation}
\label{eqn::actions}
\mathbb{A}= \{a_i | a_i = transition(s_{i-1}, s_i), i = 1, 2, ..., n\}
\end{equation}

The \emph{action} here is different from the input vector $\bm{u}_t$ defined in Eqn. \ref{eqn::discrete_state_space2}, but a more abstract representation of state transition. One example of $transition(s_{i-1}, s_i)$ is simply the difference between two states $\lVert s_i - s_{i-1} \rVert$. Exerting \emph{actions} from initial state $s_0$ will generate an ordered sequence of \emph{states}: 

\begin{equation}
\label{eqn::states}
\mathbb{S}= \{s_i | i = 0, 1, ..., n\}
\end{equation}

We further conglomerate consecutive \emph{states} starting from $s_0$ to \emph{paths}:

\begin{equation}
\label{eqn::paths}
\mathbb{P}= \{p_i | p_i \ni \{s_0, s_1, ..., s_i   \}, i = 0, 1, ..., n\}
\end{equation}

\begin{figure}[]
\centering
\includegraphics[width=1\columnwidth]{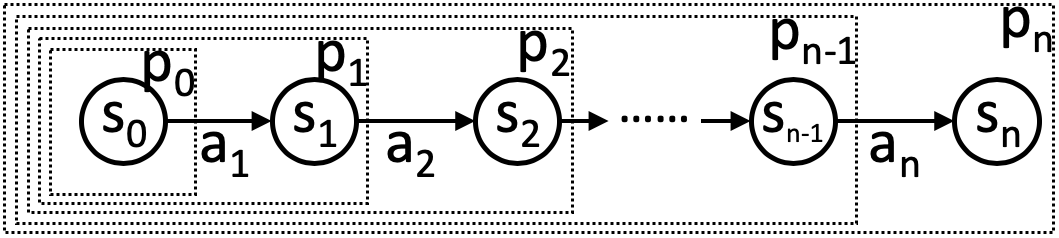}
\caption{Graphical Representation of \emph{actions}, \emph{states}, and \emph{paths}: arrows represent \emph{actions}, transitioning from previous to next \emph{state}, shown in circle. Dashed boxes represent \emph{paths}. }
\label{fig::action_state_path}
\end{figure}

A graphical representation of our augmented state space is shown in Fig. \ref{fig::action_state_path}.

\subsection{Risk Representation}
The proposed risk representation contains potential risk caused by all \emph{actions}, \emph{states}, and \emph{paths}. 

Executing any sequence of \emph{actions} can cause potential risk, regardless of which \emph{state} the robot is in and which \emph{path} it is taking. More specifically, this could be expressed as the difficulty of the \emph{action} and difference between consecutive \emph{actions} (turn). 

 \begin{equation}
\label{eqn::risk_action}
risk_a(a_1, a_2, ..., a_n) = w_{a1}\sum\limits_{i=1}^n \lVert a_i \rVert + w_{a2}\sum\limits_{i=2}^n \lVert a_i-a_{i-1} \rVert
\end{equation}

Here, $\lVert a_i \rVert $ could be viewed as the difficulty of executing $a_i$, such as the length of the \emph{action}. For example, moving in a 8-connectivity 2-D occupancy grid may have $\lVert a_i \rVert=1$ or $\lVert a_i \rVert = \sqrt{2}$. $\lVert a_i-a_{i-1} \rVert$ is the difference between consecutive \emph{actions}, such as the risk of turning or change directions. These two terms resemble the effect of first (velocity) and second (acceleration) derivatives of the original high dimensional state space. Higher order derivatives could be approximated by more terms, such as $a_i, a_{i-1}, a_{i-2}$. But for the sake of simplicity and practicality in conventional mobile robotic systems, we only aim at the risk induced by first and second derivatives. $w_{a1}$ and $w_{a2}$ are the relative weights of the two \emph{action}-dependent risk elements. 

The conventional risk representation assumes risk to be a function of \emph{state} and the function $r: s_i \mapsto ri$ is well-defined and represents all necessary risk information. The proposed explicit risk representation includes this type of risk representation, but this only composes a subset of all risk sources.

 \begin{equation}
\label{eqn::risk_states}
risk_s(s_0, s_1, ..., s_n) = \sum\limits_{i=0}^n r_s(s_i) = \sum\limits_{i=0}^n \sum\limits_{j=1}^m w_{sj}r_{sj}(s_i) 
\end{equation}

where the risk of a certain \emph{state} $r_s(\cdot)$ could be subdivided into $m$ \emph{state}-dependent risk elements $r_{sj}(\cdot)$ and $w_{sj}$ represents the respective weight. 

The final type of risk is $\emph{path}$-dependent. This kind of risk is associated with the end state of a \emph{path} and needs to be evaluated based on the entire \emph{path}:

 \begin{equation}
\label{eqn::risk_paths}
risk_p(p_0, p_1, ..., p_n) =  \sum\limits_{i=0}^n r_p(p_i) = \sum\limits_{i=0}^n \sum\limits_{j=1}^m w_{pj}r_{pj}(p_i)
\end{equation}

where $r_{pj}(\cdot)$ is one of the $m$ different \emph{path}-dependent risk elements and $w_{pj}$ is the respective weight. $r_p(p_i)$ is not the risk of executing the entire path $p_i$, but the risk associated with the end state of path $p_i$, $s_i$, by taking the path $p_i$. A more intuitive illustration is in terms of conditions, despite the abuse of conditional notation: 

\begin{equation}
\label{eqn::risk_paths_condition}
r_p(p_i) = r(s_i|s_0, s_1, ..., s_{i-1})
\end{equation}

The summation of the risk of individual \emph{paths} is to guarantee the possible decrease in \emph{path}-dependent risk will not cancel the risk already caused by previous \emph{paths}. 

Therefore, the total risk of the path is the summation of all the risk caused by \emph{actions}, \emph{states}, and \emph{paths}. Normalization or weighting may be necessary depending on different applications. 

 \begin{equation}
\label{eqn::total_risk}
risk_{total}(P) = w_{a} risk_a + w_{s} risk_s + w_{p} risk_p
\end{equation}

where $w_{a}$, $w_{s}$, $w_{p}$ are the weights assigned to \emph{action}-, \emph{state}-, and \emph{path}-dependent risk, respectively. 

\section{QUANTITATIVE EXAMPLES}
\label{sec::example}
In this section, we provide quantitative risk representations using a particular robot platform as example, a tethered Unmanned Aerial Vehicle (UAV), Fotokite Pro, which was used as a visual assistant to pair with a tele-operated Unmanned Ground Vehicle (UGV) for operations in after-disaster environments, such as Fukushima Daiichi nuclear decommisioning \cite{xiao2017visual, xiao2019autonomous}. Using the formal risk definition and explicit risk representation discussed above, examples of \emph{relative likelihood} of the UAV \emph{not being able to finish a path}, namely risk, are computed by \emph{actions}, \emph{states}, and \emph{paths}. In the examples, the planning space is represented as a 2-D occupancy grid for simplicity. Each risk element is normalized to a value between $0$ and $1$ and all weights $w$ are set to 1 so all risk elements are treated equally. 

\begin{figure}[]
\centering
\includegraphics[width=1\columnwidth]{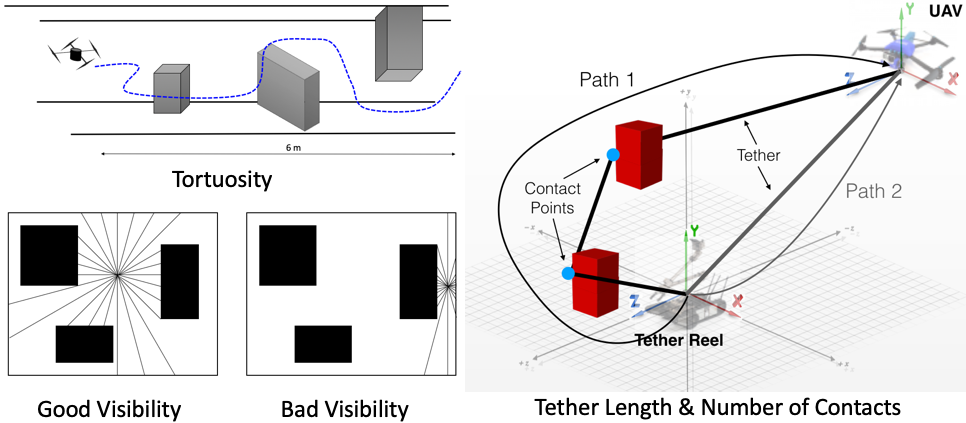}
\caption{Tortuosity (upper left), Visibility (lower left), Tether Length, and Number of Contacts (right): tortuosity captures the effort of changing \emph{action} directions. Visibility represents the obstacle density of a given \emph{state}. Tether length and number of contacts depend on the \emph{path} taken.}
\label{fig::tether_risk}
\end{figure}

Each state $s_i$ is a 2-D waypoint $(x_i, y_i)$, which we augment to $\mathbb{A}$, $\mathbb{S}$, and $\mathbb{P}$ using the approach described in Sec. \ref{sec::representation}. \emph{Action}-dependent risk is caused by \emph{action length} $\lVert a_i \rVert$ and tortuosity (Fig. \ref{fig::tether_risk} upper left) $\lVert a_i - a_{i-1} \rVert$. Absolute value of \emph{action length} is either $1$ or $\sqrt{2}$ in a 2-D occupancy grid with 8-connectivity. Then it is normalized to a value between $0$ and $1$. Tortuosity is originally defined as number of turns needed to traverse unit distance, but here it is abstracted to a high-level effort in changing action directions and normalized between $0$ and $1$. For \emph{state}-dependent risk, \emph{distance to closest obstacle} and \emph{visibiity} (Fig. \ref{fig::tether_risk} lower left) \cite{xiao2019explicit} are used. Both are normalized using membership function in fuzzy logic. The tethered UAV is a good example to illustrate \emph{path}-dependent risk. Fig. \ref{fig::tether_risk} right shows that by taking different paths, reaching the goal may have different \emph{path}-dependent risk. \cite{xiao2018motion, xiao2018indoor} show that longer \emph{tether length} and more \emph{number of contacts} in path 1 will introduce more uncertainty, and therefore more risk (\emph{state}-dependent risk for path 1 may be actually lower since the \emph{distance to closest obstacle} and \emph{visibility} are larger).  \emph{Tether length} and \emph{number of contacts} are also normalized using the potential maximum values given the map size and obstacle density at hand (here, $20$ unit length for \emph{tether length} and $10$ for \emph{number of contacts}). 

%The tethered UAV navigates from same start to same goal location with two different paths: path 1 has two contact points \cite{xiao2018motion} with a longer tether \cite{xiao2018indoor}, while path 2 does not have contact points at all and the tether is very short.

\begin{figure}
\centering
\subfloat[Path 1]{\includegraphics[width=0.5\columnwidth]{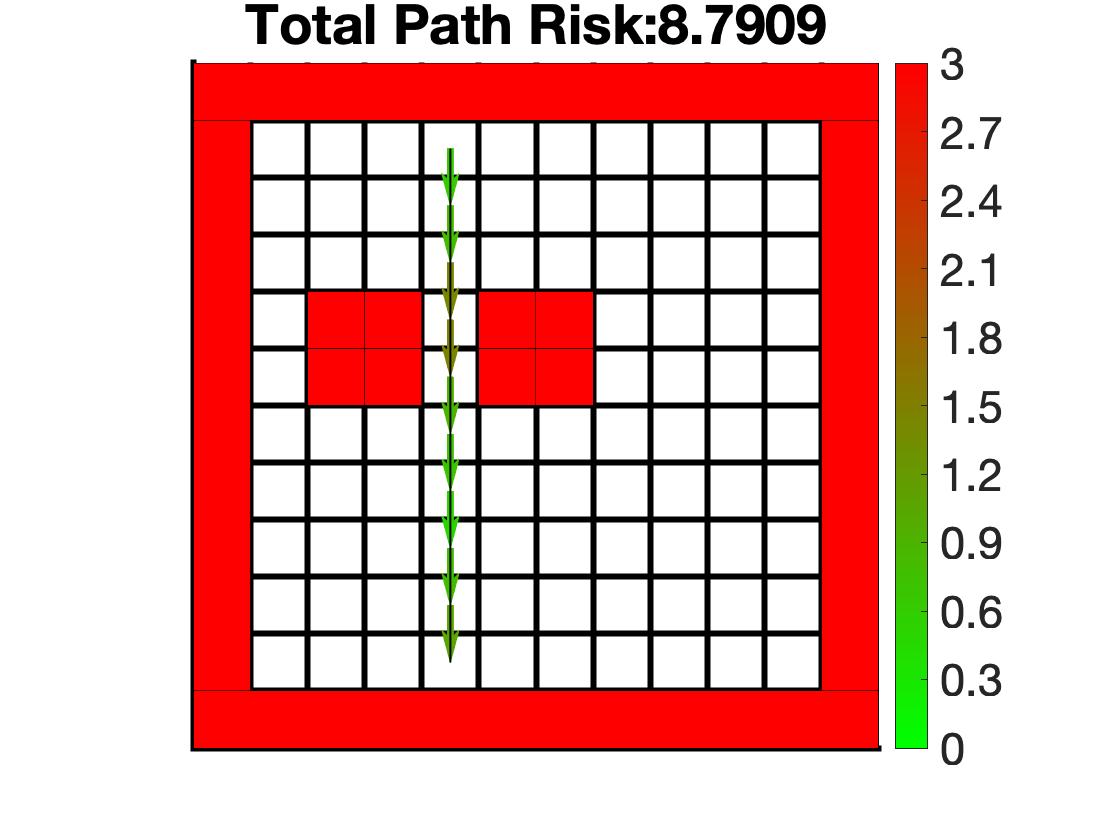}%
\label{fig::path11}}
\hfill
\subfloat[Path 2]{\includegraphics[width=0.5\columnwidth]{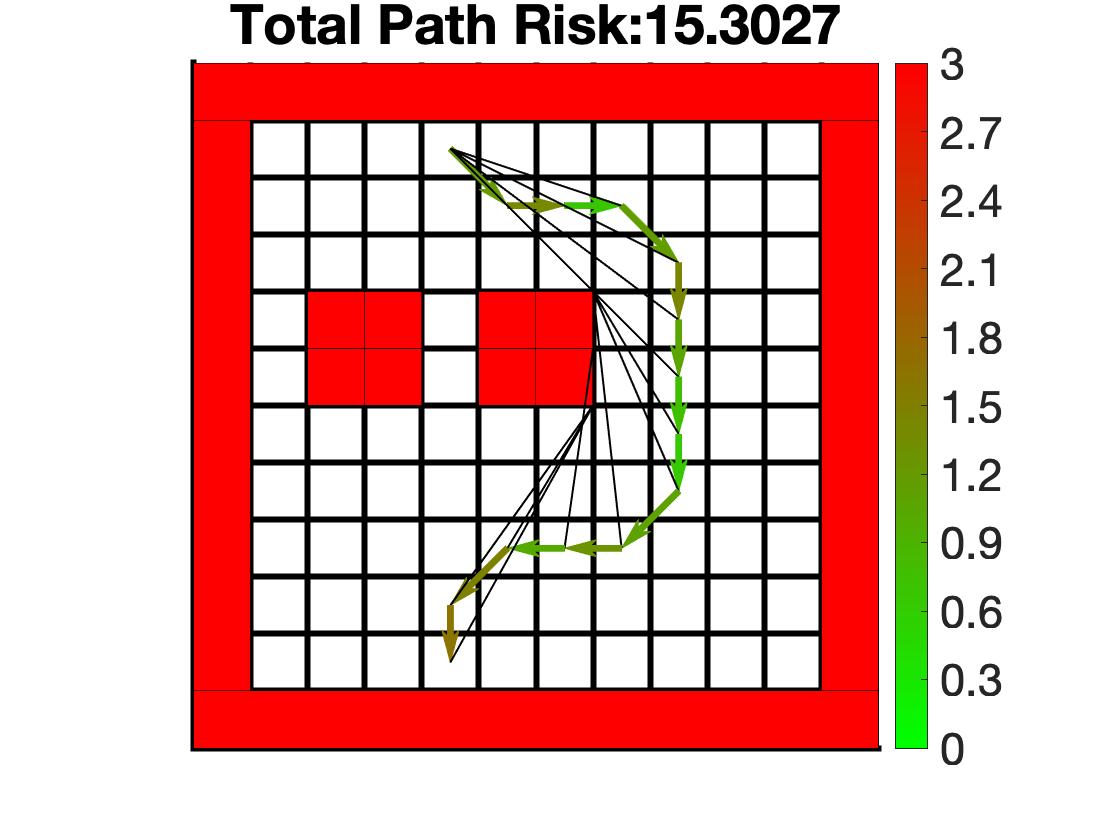}%
\label{fig::path12}}
\caption{Quantitative Risk Representation of Paths in Fig. \ref{fig::va}}
\label{fig::path1}
\end{figure}

For the two example paths shown in Fig. \ref{fig::va}, risk is quantified using the proposed risk definition and representation. In Fig. \ref{fig::path1}, red cells represent obstacles and map edges are also treated as obstacles. Arrows connect from start to goal location. Black lines denote tether configuration at each step. It is assumed that tether reel center coincides with the start location. Contacts with the obstacles could be seen as the kinks on the tether. A color map from green to red represents the risk level from low to high. Risk caused by \emph{actions}, \emph{states}, and \emph{paths} are further normalized between $0$ and $1$ again, resulting a total summation between $0$ and $3$ for each step on the path. The colors of the arrows denote the risk index of taking that step, computed by the risk representation described in Sec. \ref{sec::representation}. Due to the additivity property of the proposed risk definition, the motion risk of the entire path is the summation of all subpaths, i.e. individual steps. In Fig. \ref{fig::path11}, the straight path comes with a straight tether without any contacts. The risk caused by changes of \emph{actions} is zero. The risk is low at the very beginning, and increases due to higher \emph{state}-dependent risk when approaching the narrow gap between the two obstacles. The risk decreases after coming out of the gap. But \emph{path}-dependent risk caused by longer \emph{tether length} adds up moving away from tether reel. On the other hand (Fig. \ref{fig::path12}), the path circumvents the narrow gap between obstacles and goes through wide open spaces to approach the same goal location. Using conventional state-based risk representation, this path may be safer since most states are far away from obstacles and have higher visibility. However, extra maneuvers to go around the obstacles (\emph{actions}), longer tether, and contact points (\emph{path}) induce more risk to this path. Total risk of path 2 ($15.3$) is higher than that of path 1 ($8.8$), meaning that the relative likelihood of the robot not being able to finish path 2 is higher than that of path 1. Previously neglected aspects of risk is now considered by the proposed representation.  

%\begin{table}[]
%\centering
%\begin{tabular}{|c|c|c|c|}
%\hline
%\bf{Total Risk} & \bf{\emph{Action}-Risk} & \bf{\emph{State}-Risk} & \bf{\emph{Path}-Risk} \\ \hline \hline
%0.5000       & 0.0000        & 0.5000       & 0.0000      \\ \hline
%0.6286       & 0.3536        & 0.2500       & 0.0250      \\ \hline
%0.8000       & 0.3536        & 0.3964       & 0.0500      \\ \hline
%1.4286       & 0.3536        & 1.0000       & 0.0750      \\ \hline
%1.4536       & 0.3536        & 1.0000       & 0.1000      \\ \hline
%0.8750       & 0.3536        & 0.3964       & 0.1250      \\ \hline
%0.6945       & 0.3536        & 0.1910       & 0.1500      \\ \hline
%0.5286       & 0.3536        & 0.0000       & 0.1750      \\ \hline
%0.8036       & 0.3536        & 0.2500       & 0.2000      \\ \hline
%1.0786       & 0.3536        & 0.5000       & 0.2250      \\ \hline
%\end{tabular}
%\caption{Step Risk of Path 1}
%\end{table}

\begin{figure}
\centering
\subfloat[Path 1]{\includegraphics[width=0.5\columnwidth]{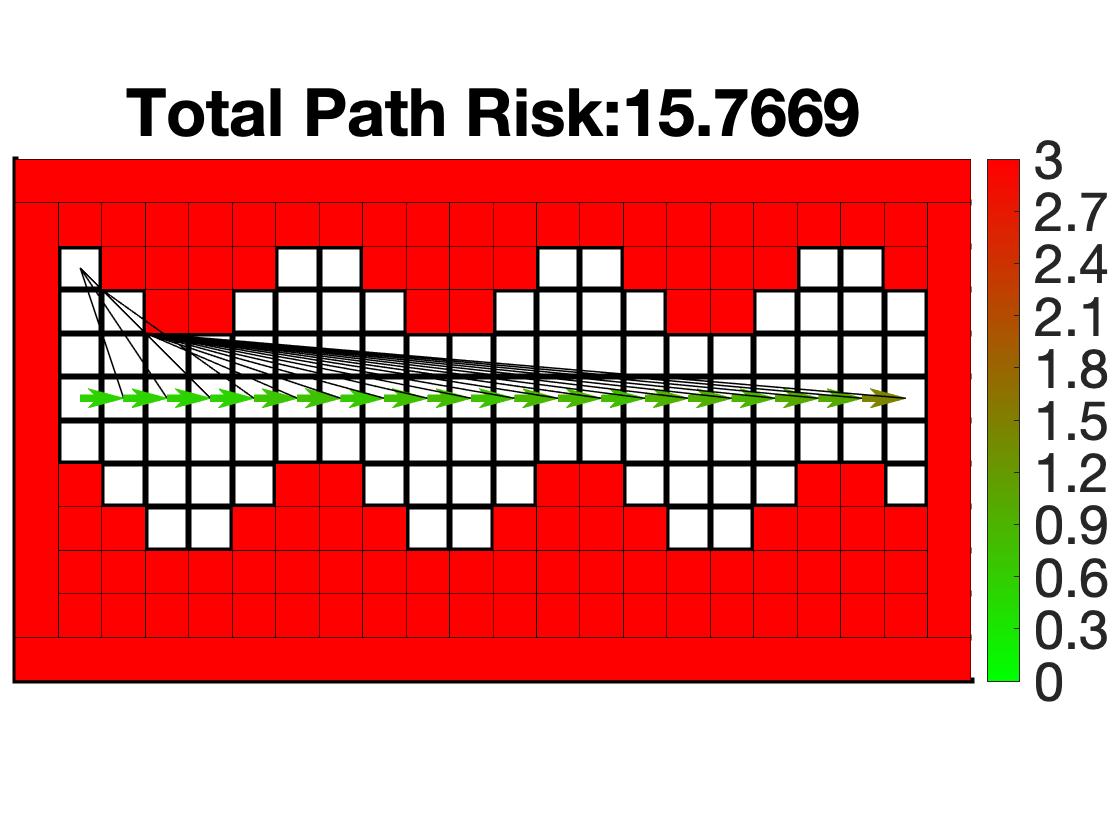}%
\label{fig::path21}}
\hfill
\subfloat[Path 2]{\includegraphics[width=0.5\columnwidth]{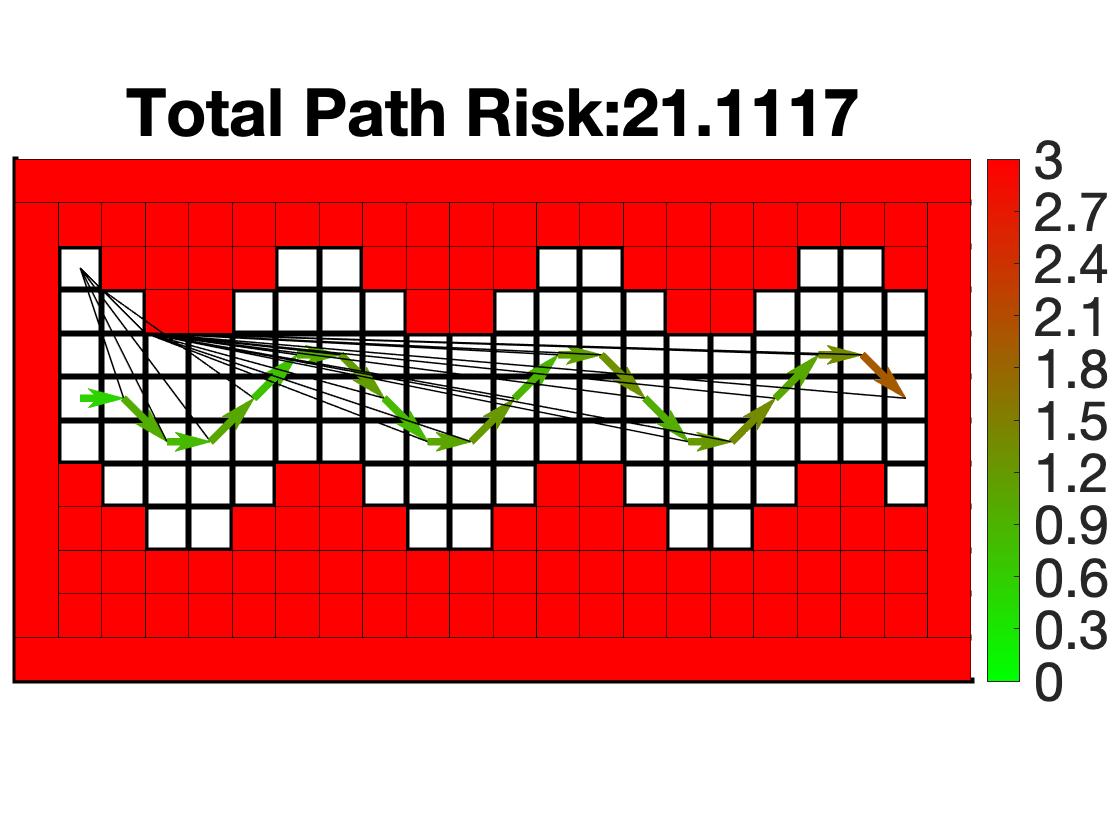}%
\label{fig::path22}}
\caption{Quantitative Risk Representation of Paths in Fig. \ref{fig::tortuous_straight}}
\label{fig::path2}
\end{figure}

Another set of examples are shown in Fig. \ref{fig::path2}, as the quantitative results of the example in Fig. \ref{fig::tortuous_straight}. Tether reel locates at the upper left corner of the map. With regular state-based approach, path 2 maneuvers through a series of ``safe'' states (as shown in the blue risk values in Fig. \ref{fig::tortuous_straight}), while states on path 1 have higher risk (orange values in Fig. \ref{fig::tortuous_straight}). However, risk caused by the \emph{actions} to maneuver the tortuous path is not considered by the traditional approach. These \emph{actions} could be risky due to frequently aggressive rotor thrusts, extra disturbances for sensing and controls, etc. \emph{Path}-level risk is also overlooked, since the tortuous path requires longer tether length on average. All these risk elements could be properly incorporated into the total motion risk by the proposed risk representation: the total risk of path 2 ($21.1$) is actually higher than that of path 1 ($15.8$). 

It is worth to note that this work does not claim to establish an unanimous measurement of risk, but only provides a formal definition and an approach to explicitly represent motion risk as a relative likelihood. In the quantitative examples, it is assumed that all weights are $1$ and all risk elements are normalized to $0$ and $1$. This treats all risk elements identically in terms of their contribution to the final path risk. No prioritization of certain risk elements exists in the examples. When using the proposed risk representation on specific robots, however, prioritization in the combination is necessary depending on the characteristic of that particular robot. Vehicle-, environment-, and mission-dependent criteria need to be considered. For example, risk of turning for holonomic vehicles (vehicle-dependent) is trivial, so importance of risk caused by difference in consecutive \emph{actions} ($w_{a2}$ in Eqn. \ref{eqn::risk_action}) could be reduced, and vice versa for nonholonomic vehicles. The vehicle's motion or perception accuracy also needs to be considered. When facing large disturbances (environment-dependent), \emph{state}-dependent risk ($w_s$ in Eqn. \ref{eqn::total_risk}) needs to be prioritized, since larger safety margin is necessary. Environment semantics, or consequences of improper interactions with the environment (for example, lethal physical damage or only deterioration of perception), also determine which risk elements should carry more weight. Safety- or stability-oriented missions (mission-dependent) play a role in prioritization as well. Therefore, we are not saying path 1  is safer in general than path 2 in Fig. \ref{fig::path1} and Fig. \ref{fig::path2}. A robot with low perception and actuation accuracy working under large disturbances may make path 2 safer in Fig. \ref{fig::path1} and Fig. \ref{fig::path2}. Our risk definition and representation provide a general framework to incorporate more relevant factors into motion risk, which is otherwise impossible with conventional state-dependent approaches only. Therefore the proposed paradigm is more comprehensive and general. 

%\section{DISCUSSIONS}
%\label{sec::discussions}

\section{CONCLUSIONS}
\label{sec::conclusions}
Robot motion risk is formally defined as the \emph{relative likelihood} of the robot \emph{not being able to finish the path}. The \emph{relative likelihood} is ordered by a numerical value, called \emph{risk index}, which is non-negative, monotonic, and additive. \emph{Not being able to finish the path} is also formally defined mathematically. Working on paths in a reduced dimensional space (2-D or 3-D) without higher order derivatives or other information, an explicit risk representation approach is presented based on augmentation of the original path in Cartesian space into \emph{actions}, \emph{states}, and \emph{paths}. Risk caused by those are represented explicitly by risk index. Examples of quantitative risk representation are provided with respect to a tethered UAV. The new representation enables the inclusion of previously impossible aspect of risk, such as risk caused by \emph{actions} and \emph{paths}, in particular, \emph{action length}, \emph{tortuosity} and \emph{tether length}, \emph{number of contacts}, respectively. Although this work does not aim at an unanimous measurement of motion risk, the proposed paradigm provides a more comprehensive and general approach to reason about motion risk. Future work will focus on applying the proposed approach to a variety of mobile robot platforms and validating the representation using physical experiments. It will also be combined with new planners in order to plan risk-aware path in a more broad sense. 

\section*{ACKNOWLEDGMENT}
This work is supported by NSF 1637955, NRI: A Collaborative Visual Assistant for Robot Operations in Unstructured or Confined Environments. 

%%%%%%%%%%%%%%%%%%%%%%%%%%%%%%%%%%%%%%%%%%%%%%%%%%%%%%%%%%%%%%%%%%%%%%%%%%%%%%%%
\bibliographystyle{IEEEtran}
\bibliography{IEEEabrv,references}

\end{document}